\documentclass{article} % For LaTeX2e
\usepackage{iclr2025_conference,times}

\usepackage{amsmath,amsfonts,bm}

\def\Figref#1{Figure~\ref{#1}}

\def\Secref#1{Section~\ref{#1}}
\def\eqref#1{equation~\ref{#1}}
\def\1{\bm{1}}

\DeclareMathAlphabet{\mathsfit}{\encodingdefault}{\sfdefault}{m}{sl}
\SetMathAlphabet{\mathsfit}{bold}{\encodingdefault}{\sfdefault}{bx}{n}

\usepackage{hyperref}
\usepackage{url}
\usepackage{graphicx}
\usepackage{xspace}
\usepackage[linesnumbered,ruled,vlined]{algorithm2e}

\usepackage{listings}
\usepackage{xcolor}

\lstdefinestyle{progressliststyle}{
    basicstyle=\ttfamily\small,
    frame=single, % A simple box frame
    frameround=tttt, % Slightly rounded corners
    breaklines=true,
    showstringspaces=false,
    moredelim=**[is][\color{green!60!black}]{@x@}{@x@},
    moredelim=**[is][\color{orange!90!black}]{@p@}{@p@},
    moredelim=**[is][\color{blue!70!black}\bfseries]{@h@}{@h@}
}

\newcommand{\LLM}{\text{LLM}\xspace}
\newcommand{\aime}{Aime\xspace}

\newcommand{\toolset}{\mathcal{T}}

\newcommand{\tool}{\tau}

\title{Aime: Towards Fully-Autonomous Multi-Agent Framework}

\author{Yexuan Shi\thanks{Corresponding Author, E-mail: shiyexuan@bytedance.com} , \quad  Mingyu Wang, \quad  Yunxiang Cao, \quad  Hongjie Lai, \quad  Junjian Lan, \\  
\textbf{Xin Han, \quad Yu Wang, \quad  Jie Geng, \quad Zhenan Li, \quad Zihao Xia, \quad Xiang Chen,} \\
\textbf{Chen Li, \quad  Jian Xu, \quad  Wenbo Duan, \quad  Yuanshuo Zhu} \\
\\
ByteDance\\
}

\begin{document}

\maketitle

\begin{abstract}
Multi-Agent Systems (MAS) powered by Large Language Models (LLMs) are emerging as a powerful paradigm for solving complex, multifaceted problems. However, the potential of these systems is often constrained by the prevalent plan-and-execute framework, which suffers from critical limitations: rigid plan execution, static agent capabilities, and inefficient communication. These weaknesses hinder their adaptability and robustness in dynamic environments. 
This paper introduces \aime{}, a novel multi-agent framework designed to overcome these challenges through dynamic, reactive planning and execution. 
\aime{} replaces the conventional static workflow with a fluid and adaptive architecture. 
Its core innovations include: (1) a Dynamic Planner that continuously refines the overall strategy based on real-time execution feedback; (2) an Actor Factory that implements Dynamic Actor instantiation, assembling specialized agents on-demand with tailored tools and knowledge; and (3) a centralized Progress Management Module that serves as a single source of truth for coherent, system-wide state awareness. 
We empirically evaluated \aime{} on a diverse suite of benchmarks spanning general reasoning (GAIA), software engineering (SWE-bench Verified), and live web navigation (WebVoyager). The results demonstrate that \aime{} consistently outperforms even highly specialized state-of-the-art agents in their respective domains. Its superior adaptability and task success rate establish \aime{} as a more resilient and effective foundation for multi-agent collaboration.

\end{abstract}

\section{Introduction}
\label{sec:intro}

The recent emergence of Large Language Models (LLMs) represents a significant milestone in artificial intelligence~\citep{DBLP:journals/tist/ChangWWWYZCYWWYZCYYX24}. 
Demonstrating profound capabilities in natural language understanding, reasoning, and generation, LLMs now serve as the foundational technology for a new class of intelligent systems. 
This trend has spurred the development of LLM Agents: autonomous entities that leverage an LLM as their central cognitive engine. 
By augmenting LLMs with external tools, such as code interpreters or information retrieval systems, these agents can interact with their environment to execute complex, multi-step tasks, acting as versatile and autonomous problem-solvers~\citep{DBLP:journals/fcsc/WangMFZYZCTCLZWW24, DBLP:conf/iclr/QinLYZYLLCTQZHT24, DBLP:journals/tmlr/WangX0MXZFA24}.

Building upon the success of individual agents, the field is increasingly exploring Multi-Agent Systems (MAS), where teams of LLM Agents collaborate to solve complex problems that exceed the capabilities of any single agent. 
This paradigm is valued for its ability to decompose large-scale problems and leverage specialized agents in a synergistic manner~\citep{DBLP:conf/nips/0001ST00Z23, DBLP:conf/iclr/HongZCZCWZWYLZR24, DBLP:conf/nips/0003ZWZ0C24}.
Among the various architectures, the \textit{plan-and-execute} framework has become a dominant approach~\citep{DBLP:journals/corr/abs-2402-02716}. 
In this structure, a dedicated planner agent deconstructs a user's request into a static sequence of subtasks. These subtasks are then assigned to a team of executor agents, each possessing a predefined role and toolset.

Despite its widespread adoption, the plan-and-execute model exhibits critical limitations, particularly in dynamic or unpredictable environments. 
The rigid separation of planning and execution introduces significant operational friction. 
Our work identifies three fundamental challenges that curtail the effectiveness of these systems:
    (1) \textit{Rigid Plan Execution.} Plans are generated once and are typically brittle. The planner remains idle during execution, rendering the system unable to adapt to real-time feedback or unexpected outcomes produced by the executors.
    (2) \textit{Static Agent Capabilities.} Agents are confined to predefined roles and toolkits. This rigidity limits the system's ability to handle unforeseen tasks that demand novel skills, thereby compromising its extensibility and robustness.
    (3) \textit{Inefficient Communication.} Task handoffs between agents often result in context loss. Without a centralized state management system, agents operate with an incomplete view of the overall progress, leading to redundant work and coordination failures.

To address these limitations, this paper introduce \aime{}, a novel multi-agent framework designed for dynamic, reactive planning and execution. Instead of a static, top-down workflow, \aime{} operates on a principle of continuous adaptation. It replaces the fixed planner-executor dichotomy with a more fluid architecture comprising four core components: a \textit{Dynamic Planner} that continuously refines strategy based on live feedback; an \textit{Actor Factory} that instantiates specialized actors on-demand to meet specific task requirements; autonomous \textit{Dynamic Actors} that execute tasks; and a centralized \textit{Progress Management Module} that serves as a single source of truth for system-wide state awareness.

The primary contributions of this work are as follows:
\begin{itemize}
    \item We propose \aime{}, a novel MAS framework that replaces the rigid plan-and-execute paradigm with a fluid, adaptive system. \aime{} enables dynamic plan adjustments, on-demand role allocation, and streamlined coordination to effectively manage complex, evolving tasks.
    
    \item We introduce \textit{Dynamic Actor Instantiation}, a mechanism implemented via an \textit{Actor Factory}. 
    This component assembles specialized actors on-demand, equipping them with tailored personas, tools, and knowledge to address the limitations of static agent roles.

    \item We design a centralized \textit{Progress Management Module} that maintains a unified, real-time view of task progress. This module mitigates information loss and coordination failures by providing coherent state awareness across the entire system.

    \item We conduct comprehensive experiments on challenging benchmarks for general reasoning, software engineering, and web navigation (GAIA, SWE-bench, WebVoyager). Our results show that it significantly outperforms specialized state-of-the-art frameworks in both task success rate and adaptability.
\end{itemize}

The remainder of this paper is organized as follows. \Secref{sec:background} provides a detailed background on LLM Agents and Multi-Agent Systems. \Secref{sec:overview} presents a high-level overview of the \aime{} framework. \Secref{sec:method} details the design of its core components. Experimental setup and results are presented in \Secref{sec:exp}. We then discuss the related work in \Secref{sec:related}, and finally, conclude the paper in \Secref{sec:conclusion}.
\section{Background}
\label{sec:background}

This section reviews the core concepts of Large Language Model Agents (LLM Agents) and Multi-Agent Systems (MAS). We then discuss current challenges and present our research motivation.

\paragraph{LLM Agent.}

The emergence of Large Language Models (LLMs) has enabled a new class of autonomous systems known as LLM Agents.
An LLM Agent utilizes an LLM not merely as a text generator, but as its central cognitive core for reasoning, planning, and decision-making~\citep{DBLP:conf/iclr/YaoZYDSN023}.
To transcend the inherent limitations of their static, pre-trained knowledge, these agents are augmented with external tools.
This augmentation allows them to interact dynamically with their environment, granting them capabilities such as executing code, retrieving real-time information from the web, and controlling external devices~\citep{DBLP:conf/www/ShiGY0CCYVR25}.
Formally, an LLM Agent, $A$, can be defined as a tuple:

\begin{equation*}
A = \{\LLM, \mathcal{T}, P, M\}
\end{equation*}

where $\LLM$ is the cognitive engine; $\toolset = \{\tool_1, \ldots, \tool_n\}$ is a set of external tools, each with well-defined functionalities; $P$ represents the prompts that structure the LLM's reasoning process and tool-use strategies; and $M$ is a memory module for storing historical interactions, states, and contextual information~\citep{DBLP:journals/corr/abs-2404-13501}.

The operational paradigm of an LLM Agent is typically an iterative cycle of reasoning, acting, and observing.
The agent assesses a given goal, formulates a plan, executes an action (often by invoking a tool), and then incorporates the resulting feedback to inform its next step.
This loop continues until the task is accomplished, enabling LLM Agents to deconstruct and solve complex, multi-step problems.

\paragraph{Multi-Agent System.}
Building upon the capabilities of individual LLM Agents, Multi-Agent Systems (MAS) represent the next frontier in collaborative AI.
A MAS is composed of multiple autonomous agents operating within a shared environment, orchestrated to tackle objectives that are too complex or large in scope for a single agent to handle effectively~\citep{DBLP:conf/ijcai/GuoCWCPCW024}. 
Through structured interaction and role specialization, these agents can achieve emergent intelligence and synergy.
A MAS can be formally described as:

\begin{equation*}
MAS = S(A_1, A_2, \cdots, A_m)
\end{equation*}

where each $A_i$ is an LLM Agent as previously defined, and $S$ denotes the collaboration framework that governs their interactions.
This framework defines agent roles, communication protocols, and potential hierarchies.

A defining characteristic of LLM-based MAS is the use of natural language as the primary medium for inter-agent communication.
This offers unprecedented flexibility compared to traditional, rigidly coded protocols.
However, it also introduces significant challenges in maintaining semantic consistency, ensuring goal alignment, and managing complex conversational flows.
The design of an effective collaboration structure $S$ is therefore critical to system performance and remains an active area of research, which leads to the challenges we address in this work.

\paragraph{Motivation.}
In current MAS research and applications, a widely adopted orchestrated structure $S$ is the plan-and-execute framework~\citep{DBLP:journals/corr/abs-2402-02716}. 
This framework assigns specialized roles to agents, typically as planners or executors. The workflow generally unfolds in three distinct stages: 
(1) \textit{Global Planning}, where a planner agent analyzes a request and decomposes it into structured subtasks; 
(2) \textit{Task Assignment}, where subtasks are allocated to executor agents based on their predefined capabilities; and 
(3) \textit{Execution and Feedback}, where executors complete their assigned tasks and report the outcomes.

While this paradigm offers a structured approach to multi-agent collaboration, it struggles with complex, dynamic tasks.
The strict separation of planning and execution often leads to suboptimal performance, particularly in environments where task requirements can change mid-execution.
Our work is motivated by three critical challenges inherent in this model:

\begin{itemize}
    \item{\textit{Rigid Plan Execution.}} 
    Conventional plans are often static. Once formulated, the planner must typically wait for all executors to report completion before aggregating results. However, the autonomy of individual executors means their actions may deviate from the prescribed plan, leading to incomplete or redundant work. This deviation provides the planner with unreliable feedback, ultimately degrading overall system performance~\citep{DBLP:journals/corr/abs-2503-13657}.
    
    \item{\textit{Static Agent Capabilities.}}
    The plan-and-execute model presupposes that the predefined capabilities of agents are sufficient for all anticipated tasks. This assumption is often violated in practice. For instance, inaccurate or incomplete descriptions of an agent's skills can lead the planner to make suboptimal task assignments~\citep{DBLP:conf/nips/0001ST00Z23}. Moreover, the system cannot adapt to unforeseen tasks that demand new tools or capabilities, thus limiting its extensibility and performance on novel problems.
    
    \item{\textit{Inefficient Communication.}}
    The handoff of information between agents is frequently a bottleneck. Critical context can be lost or distorted during task delegation, hindering seamless execution~\citep{cognition2025dontbuildmultiagents}. This issue is exacerbated by the absence of a shared state management system. Agents often operate with an incomplete view of the overall progress, as status updates are typically aggregated only upon task completion. This lack of real-time shared awareness can lead to redundant efforts and critical delays.
\end{itemize}

These challenges collectively limit the effectiveness and adaptability of existing multi-agent systems~\citep{DBLP:journals/corr/abs-2503-13657}. Overcoming these limitations requires a more advanced framework that enables flexible execution, dynamic agent roles, and reliable, context-aware communication.

\section{\aime Overview}
\label{sec:overview}

\aime is a novel multi-agent system that transforms the traditional static plan-and-execute paradigm into dynamic, reactive planning and execution. 
It operates on the principle of dynamic adaptation, where both task allocation and agent capabilities evolve based on real-time execution feedback and progress updates.

\subsection{Framework}
\label{subsec:overview}

The \aime framework consists of four core components that work in concert to enable dynamic multi-agent collaboration.

\begin{figure}[h]
\begin{center}
\includegraphics[width=\linewidth]{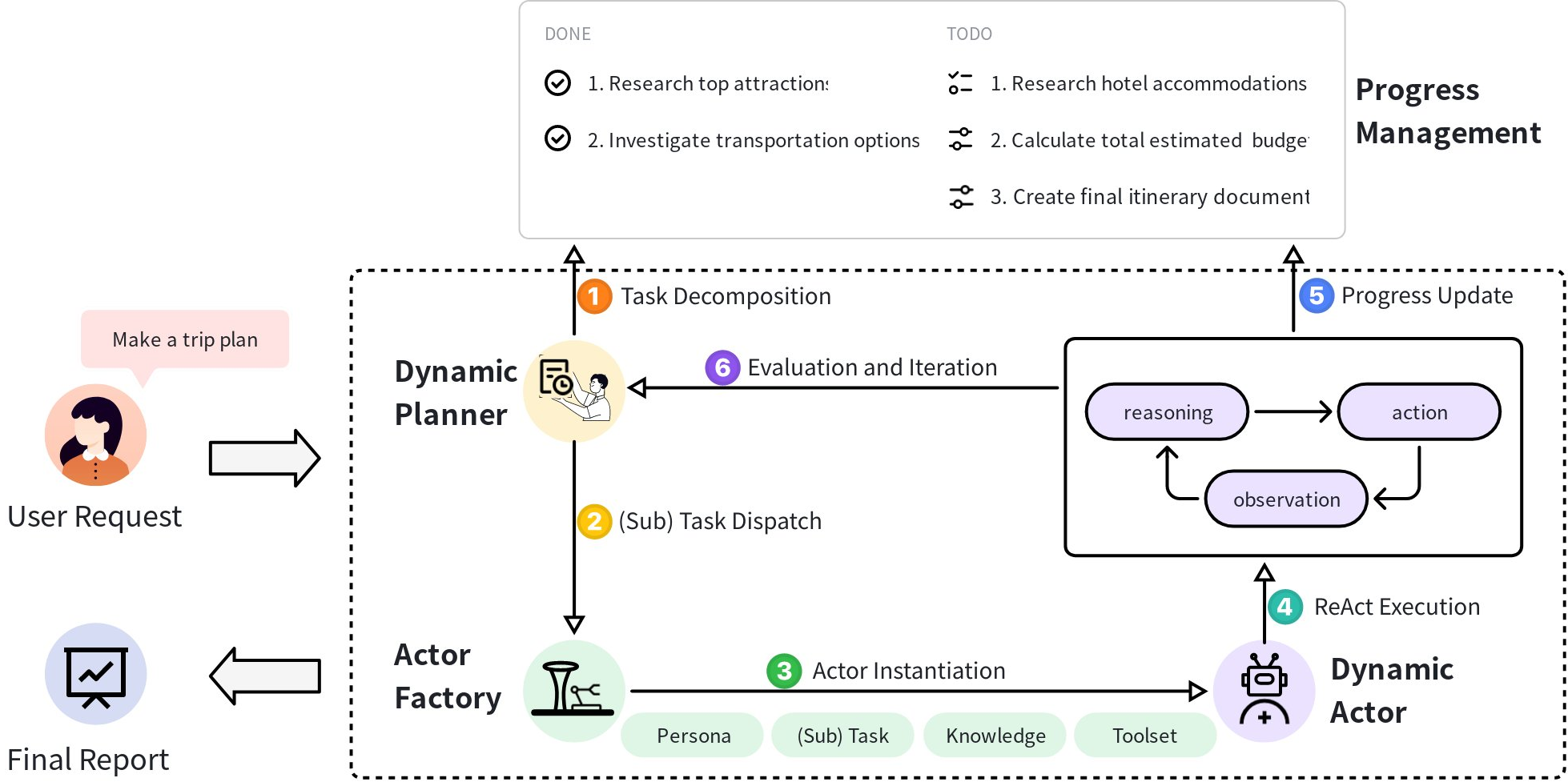}
\end{center}
\caption{The workflow of \aime framework.}
\label{fig:framework}
\end{figure}

\paragraph{Dynamic Planner.}
The \textit{Dynamic Planner} serves as the central orchestrator for task management.
It decomposes high-level objectives into a hierarchical structure of executable subtasks and maintains a global task list that tracks the status of each subtask. 
It continuously monitors execution progress and dynamically adapts the plan based on feedback from \textit{Dynamic Actors} and status updates from the \textit{Progress Management Module}.

\paragraph{Actor Factory.} 
The \textit{Actor Factory} is responsible for instantiating specialized actors tailored to specific subtask requirements.
Upon receiving a subtask, the factory analyzes its specifications to determine the optimal configuration for an actor.
This process involves selecting an appropriate persona, a relevant knowledge base, and a necessary set of tools.
The factory then assembles the actor with customized prompts and system configurations, ensuring each actor is purpose-built for its assigned task.

\paragraph{Dynamic Actor.}
A \textit{Dynamic Actor} is an autonomous agent that executes specific subtasks assigned by the \textit{Dynamic Planner}.
Each actor employs the ReAct framework~\citep{DBLP:conf/iclr/YaoZYDSN023}, operating through iterative cycles of ``Reasoning" and ``Action".
Within each cycle, the actor selects the most suitable tool from its pre-configured toolkit, performs an action, and then evaluates the outcome based on the resulting observation.
This loop continues until the subtask's completion criteria are met.

\paragraph{Progress Management Module.}
The \textit{Progress Management Module} functions as the shared memory and central state for system-wide coordination.
It maintains a structured representation of the entire task hierarchy and the real-time status of all subtasks.
This centralized record ensures a consistent understanding of progress among the \textbf{Dynamic Planner} and all actors.
Crucially, by embedding explicit completion criteria within each task entry, the module provides clear and objective standards for validating task completion.

\subsection{Workflow}
\label{subsec:workflow}
The \aime framework operates through the iterative workflow that orchestrates the dynamic collaboration among its components.
As illustrated in \Figref{fig:framework}, the process begins with a user request and proceeds through a cycle of planning, actor instantiation, execution, and state updates, detailed as follows:

\begin{description}
    \item[Step 1: Task Decomposition.] 
    The workflow commences when the \textit{Dynamic Planner} receives a task from the user.
    The planner decomposes this request into a structured plan of subtasks and initializes the corresponding task list in the \textit{Progress Management Module}.

    \item[Step 2: (Sub)Task Dispatch.] 
    The \textit{Dynamic Planner} then identifies the next executable subtask from the plan and dispatches its specification to the \textit{Actor Factory}.

     \item[Step 3: Actor Instantiation.] 
    Upon receiving the subtask specification, the \textit{Actor Factory} instantiates a specialized \textit{Dynamic Actor}.
    This actor is purpose-built for the subtask, equipped with the precise persona, knowledge, and tools required for effective execution.

    \item[Step 4: ReAct Execution.] 
    The newly instantiated actor executes its assigned subtask by following the ReAct paradigm.
    It operates in a loop of reasoning and action, invoking tools from its toolkit to make incremental progress toward the subtask's goal.

    \item[Step 5: Progress Update.] 
    During execution, the actor continuously reports status updates to the \textit{Progress Management Module}. This ensures that the global task state remains synchronized and accessible to all components, particularly the \textit{Dynamic Planner}.

    \item[Step 6: Evaluation and Iteration.] 
    Once a subtask is completed, the actor reports the final outcome to the \textit{Dynamic Planner}. 
    The planner evaluates this outcome, updates the global plan, and returns to Step 2 to dispatch the next available subtask.
    This cycle repeats until the top-level user request is successfully fulfilled.
\end{description}

This integrated workflow directly addresses several key challenges in multi-agent collaboration. 
The dynamic planning and dispatch loop (Steps 2 \& 6) ensures context-aware task allocation, overcoming the rigidity of static, predefined plans.
The centralized \textit{Progress Management Module} (Step 1\& 5) provides a single source of truth for task status, ensuring efficient information sharing and reducing communication overhead. 
Finally, on-the-fly actor instantiation (Step 3) allows for flexible role definition, creating actors precisely tailored to the task at hand, unlike systems with fixed actor roles.

\section{Methodology}
\label{sec:method}

This section details the core components of the \aime{} framework: the \textit{Dynamic Planner} (\Secref{subsec:planner}), the \textit{Actor Factory} (\Secref{subsec:factory}), and the \textit{Dynamic Actor} (\Secref{subsec:actor}). We then explain how these components interact via the \textit{Progress Management Module} (\Secref{subsec:progress}), which serves as the central coordination hub.

\subsection{Dynamic Planner}
\label{subsec:planner}

The \textit{Dynamic Planner} is designed to address the execution rigidity inherent in traditional plan-and-execute frameworks.
In such frameworks, the planner typically remains idle until all subtasks are complete, resulting in a bottleneck for adaptation.
In contrast, our \textit{Dynamic Planner} integrates a high-level strategic overview with adaptive, incremental execution.
This dual focus allows the system to maintain a clear path toward the final goal while remaining responsive to execution dynamics.

At its core, the planner manages two operational levels: maintaining the global task structure and determining the immediate next action.
We formalize this dual responsibility as a single, iterative reasoning step.
Given an overall goal $G$, at each step $t$, the planner, denoted as agent $A_{\text{planner}}$, assesses the current task list, $\mathcal{L}_t$, and the history of past outcomes, $\mathcal{H}_t = \{o_1, \ldots, o_{t}\}$.
Its operation is defined by the function:

\begin{equation}
(\mathcal{L}_{t+1}, g_{t+1}) = \LLM_{\text{planner}}(P_{\text{planner}}, (G, \mathcal{L}_t, \mathcal{H}_t))
\label{eq:planner}
\end{equation}

The planner produces two outputs in each iteration:
\begin{itemize}
    \item $\mathcal{L}_{t+1}$ is the updated global task list. This represents the planner's \textit{strategic} or "big-step" output, reflecting a revised understanding of the task hierarchy based on new information. If no major strategic change is required, $\mathcal{L}_{t+1}$ may simply be a copy of $\mathcal{L}_t$ with updated task statuses.
    \item $g_{t+1}$ is the specific, executable action for the system to perform next. This is the \textit{tactical} or "small-step" output, such as dispatching a subtask to the \textit{Actor Factory}.
\end{itemize}

This formulation enables remarkable adaptability.
For instance, if a subtask fails, the planner can make both strategic and tactical adjustments in a single iteration.
Its strategic reasoning might modify the global plan $\mathcal{L}_{t+1}$ to include a new contingency subtask.
Concurrently, its tactical decision $g_{t+1}$ would be to dispatch this new subtask for immediate execution.
This seamless integration of planning and re-planning is a key advantage of our approach.

The planner's interaction with the system state is strictly mediated through the structured task list $\mathcal{L}$ maintained by the \textit{Progress Management Module}.
This disciplined approach ensures state consistency across the system.
By maintaining a dynamic equilibrium between global planning and immediate action, the \textit{Dynamic Planner} serves as the cornerstone of \aime's resilience and effectiveness in complex, evolving environments.

\subsection{Actor Factory}
\label{subsec:factory}

To overcome the limitations of predefined agent roles, \aime{} introduces the \textit{Actor Factory}, a component that implements what we term \textit{Dynamic Actor Instantiation}.
Instead of selecting from a fixed pool of general-purpose agents, the factory assembles specialized actors on-demand, tailored to the precise requirements of a given subtask $g_{t}$.

Upon receiving subtask $g_{t}$ from the \textit{Dynamic Planner}, the factory analyzes its specifications to determine the necessary capabilities.
It then constructs a new actor, $A_t$, by selecting and composing components from curated pools.
This generation process is defined as:

\begin{equation}
A_t = \mathcal{F}_{\text{factory}}(g_t)
\quad \text{where} \quad
A_t = \{\LLM_t, \toolset_t, P_t, M_t\}
\end{equation}

The factory's primary functions are to select a dedicated toolkit $\toolset_t$ and construct a customized system prompt $P_t$.

\paragraph{Toolkit Selection.}
A significant challenge in complex MAS is managing a large and diverse set of tools.
Presenting all available tools to an agent's LLM can lead to inefficient selection or errors.
To mitigate this, \aime{} organizes tools into pre-packaged bundles, each catering to a specific functional category (e.g., `WebSearch' bundle, `FileSystem' bundle).
The factory selects appropriate bundles to form a final toolkit, $\toolset_t$, rather than picking from a flat list of individual tools. 
This bundle-based approach ensures functional completeness and reduces the risk of critical tool omissions.

\paragraph{Prompt Generation.}
The system prompt $P_t$ is dynamically assembled from several modular components to create a precise operational context for the actor:

\begin{equation}
P_t = \text{Compose}(\rho_t, \text{desc}(\toolset_{t}), \kappa_t, \varepsilon, \Gamma)
\end{equation}

where the components are:
\begin{itemize}
    \item{\textbf{Persona ($\rho_t$).}} Defines the actor's professional role and expertise (e.g., ``An expert travel planner specializing in creating unique and memorable journeys.''). The persona is generated to align with the subtask $g_t$, effectively creating a dedicated expert.
    \item{\textbf{Tool Descriptions ($\text{desc}(\toolset_t)$).}} Provides a concise, textual description of the selected toolkit $\toolset_t$. Supplying a minimal yet sufficient set of tools narrows the LLM's decision space, improving focus and performance.
    \item{\textbf{Knowledge ($\kappa_t$).}} Consists of highly relevant information dynamically retrieved from a knowledge base to support the subtask. For a trip-planning task, this could include tips on identifying local attractions.
    \item{\textbf{Environment ($\varepsilon$).}} Provides global context, such as operating system details or system-wide constraints (e.g., current time, access permissions), ensuring the actor's actions are environmentally aware.
    \item{\textbf{Format ($\Gamma$).}} Specifies the required output structure (e.g., a JSON schema), ensuring that the actor's responses can be reliably parsed for automated processing and state updates.
\end{itemize}

\paragraph{Summary.}
This on-the-fly instantiation mechanism offers two distinct advantages over systems with static agent roles.
First, it equips actors with the exact capabilities required for the task, eliminating both capability gaps and the cognitive load of irrelevant tools.
Second, it enhances system extensibility; new capabilities can be introduced by simply adding new tool bundles or knowledge modules, without the costly process of redesigning and re-validating a large set of static agent archetypes.

% By this on-the-fly instantiation mechanism, actors are equipped with the exact capabilities required for the task. This eliminates both capability gaps, where an agent lacks a needed skill, and superfluous complexity, where an agent is burdened with irrelevant tools. We further reduce the cognitive load on the LLM. by providing a minimal, pre-packaged toolkit instead of a vast collection of individual tools. It is the most distinct advantages over systems with static agent roles.

% On the other hand, the system can be easily extended. New capabilities can be introduced by simply adding new tool bundles or knowledge modules, without the costly process of redefining and testing a large set of static agent archetypes.

\subsection{Dynamic Actor}
\label{subsec:actor}

Once instantiated by the \textit{Actor Factory}, a \textit{Dynamic Actor} functions as an autonomous agent dedicated to executing its assigned subtask, $g_t$.
Its behavior is governed by the ReAct paradigm~\citep{DBLP:conf/iclr/YaoZYDSN023}, which integrates reasoning and action into an iterative execution cycle.

The actor, $A_t$, executes its subtask by repeatedly invoking its core LLM.
At each step $k$ of its internal loop, the actor analyzes its objective and local history to generate a new thought ($\text{thought}_{k+1}$) and a subsequent action ($\text{action}_{k+1}$).
This process is formalized as:

\begin{equation}
(\text{thought}_{k+1}, \text{action}_{k+1}) = \LLM_t(P_t, (g_t, \mathcal{H}_k))
\end{equation}

where $\mathcal{H}_k$ is the sequence of previous (action, observation) pairs stored in the actor's local memory $M_t$.
This cycle unfolds through three distinct phases:

\begin{itemize}
    \item \textbf{Reasoning.} The actor reflects on the subtask goal, its past actions, and the resulting observations to formulate a plan for the next immediate step.
    
     \item \textbf{Action.} Based on its reasoning, the actor selects and executes an action, typically a call to a tool from its specialized toolkit $\toolset_t$.

    \item \textbf{Observation.} The actor receives the output from the executed tool. This new observation is appended to its history $\mathcal{H}_k$ and serves as critical context for the next reasoning phase.
\end{itemize}

A key feature of the \textit{Dynamic Actor} is its ability to communicate progress proactively.
To facilitate this, the actor's toolkit $\toolset_t$ is augmented with a special system-provided tool: $\texttt{Update\_Progress(status, message)}$. 
Crucially, the decision to invoke this tool is not hard-coded; rather, the actor's LLM autonomously determines the appropriate moments for reporting, such as after completing a significant milestone or encountering an obstacle.
This mechanism provides the \textit{Dynamic Planner} with a near real-time view of ongoing activities without interrupting the actor's primary workflow.

The execution loop terminates when the subtask's completion criteria are met.
At this point, the actor generates a final, structured report $o_t$ for the \textit{Dynamic Planner}.
This report includes a conclusive summary of the outcome, a final status update, and any relevant artifacts (e.g., file paths, data outputs) required by subsequent tasks.

\subsection{Progress Management Module}
\label{subsec:progress}

A central challenge in multi-agent systems is maintaining a coherent and globally consistent understanding of task progress.
The \textit{Progress Management Module} addresses this by serving as the framework's centralized state manager, establishing a \textit{single source of truth} for the entire task hierarchy.
This ensures that the \textit{Dynamic Planner} and all \textit{Dynamic Actors} operate on a shared, unified view of the system's state.

\subsubsection{Core Data Structure: The Progress List}

The cornerstone of this module is a globally accessible, hierarchical data structure we call the \textbf{progress list}, denoted by $\mathcal{L}$.
It represents the complete task decomposition, from high-level objectives to granular subtasks.
A typical implementation uses a human-readable and machine-parsable format like a Markdown task list:

\begin{lstlisting}[style=progressliststyle]
- @h@Objective 1@h@: Perform Initial Research
    - @x@[x]@x@ Sub-objective 1.1: Research top attractions
    - @x@[x]@x@ Sub-objective 1.2: Investigate transportation options
- @h@Objective 2@h@: Finalize Itinerary and Budget
    - @p@[ ]@p@ Sub-objective 2.1: Research hotel accommodations
    - @p@[ ]@p@ Sub-objective 2.2: Calculate total estimated budget
    - @p@[ ]@p@ Sub-objective 2.3: Create final itinerary document
\end{lstlisting}

The key characteristics of the progress list are:
\begin{itemize}
    \item \textbf{Real-time Status Tracking.} Each item is marked with its current status (e.g., completed `[x]', pending `[ ]'), providing an at-a-glance view of system-wide progress.
    
    \item \textbf{Embedded Context and Dependencies.} The hierarchical structure implicitly encodes dependencies between tasks. Furthermore, each item can embed or link to explicit completion criteria, providing objective standards for validation.
\end{itemize}

\subsubsection{Coordination via Progress Updates}

Coordination between the planner and actors is achieved through two communication protocols targeting the progress list: real-time synchronization during execution and structured conclusion upon task completion.

\paragraph{Real-time Synchronization.}
As detailed in \Secref{subsec:actor}, each \textit{Dynamic Actor} can autonomously report incremental progress by invoking the $\texttt{Update\_Progress}$ tool.
This action pushes updates to the progress list, allowing the actor to signal key milestones (e.g., "shortlisted three potential hotels in Tokyo") or flag issues (e.g., "direct flights on the desired date are fully booked") before the entire subtask is finished.
This mechanism provides the \textit{Dynamic Planner} with high-fidelity, near real-time visibility into ongoing activities, enabling more proactive and informed decision-making.

\paragraph{Structured Task Conclusion.}
When an actor $A_t$ completes its assigned subtask, it communicates the final outcome to the \textit{Dynamic Planner} using a standardized conclusion report, $o_t$.
This message triggers a formal update to the global state, which the planner uses to modify the progress list. 

The final report $o_t$ is a structured payload composed of three essential parts:
\begin{itemize}
    \item{\textbf{Status Update.}} An explicit update for the assigned items in the progress list, marking them as completed or failed.

    \item{\textbf{Conclusion Summary.}} A narrative summary of the task's execution. This includes the final outcome, obstacles encountered, and key insights, providing rich context beyond a simple success/fail flag.

    \item{\textbf{Reference Pointers.}} A structured collection of pointers to critical artifacts produced during the task (e.g., files, database record IDs, URLs), ensuring that outputs are traceable and accessible for subsequent tasks.
\end{itemize}

By combining a shared data structure with dual communication protocols—one for real-time updates and one for final conclusions—the \textit{Progress Management Module} ensures that context is explicitly maintained, accurately updated, and efficiently transferred throughout the task lifecycle. This forms a robust foundation for dynamic multi-agent collaboration.

\section{Experiments}
\label{sec:exp}

To evaluate the effectiveness of our proposed framework, we conducted a series of experiments across three diverse and challenging benchmarks. 

% Our evaluation is designed to answer three primary questions: 

% \textbf{RQ1.} How does \aime{} perform against state-of-the-art specialized agent frameworks? 

% \textbf{RQ2.} How robust is \aime{} across different domains, from general reasoning to software engineering and web navigation? 

% \textbf{RQ3.} What is the specific contribution of each core component within the \aime{} framework?

\subsection{Experimental Setup}
\label{subsec:setup}

\paragraph{Datasets.}
We selected three benchmarks that represent a broad spectrum of complex, multi-step tasks for autonomous agents:
\begin{itemize}
    \item \textbf{GAIA}~\citep{DBLP:conf/iclr/MialonF0LS24} is a challenging benchmark for general AI assistants, comprising questions that require multi-step reasoning, tool use, and comprehension of multi-modal content. We evaluate on the public test set using the official exact string matching metric.
    
    \item \textbf{SWE-bench Verified}~\citep{DBLP:conf/iclr/JimenezYWYPPN24} is a curated subset of SWE-bench for assessing an agent's ability to resolve real-world software engineering problems. Success is rigorously evaluated by running unit tests to ensure the provided fix is correct and introduces no regressions.
    
    \item \textbf{WebVoyager}~\citep{DBLP:conf/acl/HeYM0D0L024} is an end-to-end benchmark for web agents that interact with live websites. Performance is measured by task success rate on 15 real-world sites.
\end{itemize}

\paragraph{Baselines.}
We compare \aime{} against state-of-the-art specialized baselines for each domain. To ensure a fair comparison, all agents, including our own, are powered by the same underlying LLM, where applicable.
\begin{itemize}
    \item On GAIA, we compare against leading general-purpose agent frameworks: Langfun, Trase, and OWL~\citep{DBLP:journals/corr/abs-2505-23885}.
    \item On SWE-bench Verified, our baselines are top-performing code agents: SWE-agent~\citep{DBLP:conf/nips/YangJWLYNP24} and OpenHands~\citep{DBLP:conf/iclr/0001LSXTZPSLSTL25}.
    \item On WebVoyager, we compare against prominent web agents: Browser use~\citep{browser_use2024}, Operator~\citep{openai2025operator}, and Skyvern~\citep{skyvern2025}.
\end{itemize}

\subsection{Experimental Results}
\label{subsec:overall_eval}

Table~\ref{tab:main_results} presents the main experimental results, comparing \aime{} against state-of-the-art specialized agents on their respective benchmarks. The data clearly demonstrates that our framework not only competes with but consistently outperforms these highly-tuned systems, establishing its strong generalization capabilities and a new state-of-the-art across these diverse domains.

\begin{table}[t]
\caption{Performance comparison of \aime{} against specialized baselines across three benchmarks. Baselines are evaluated only on their target domain, while \aime{} is evaluated on all three. Best scores in each column are in \textbf{bold}.}
\label{tab:main_results}
\begin{center}
\begin{tabular}{lccc}
\multicolumn{1}{c}{\bf Model}  &\multicolumn{1}{c}{\bf GAIA} &\multicolumn{1}{c}{\bf SWE-Bench Verified} &\multicolumn{1}{c}{\bf WebVoyager}
\\ 
& \small{(Success Rate \%)} & \small{(Resolved \%)} & \small{(Success Rate \%)} \\
\hline \\
\multicolumn{1}{l}{\textit{General-Purpose Agents}} \\
\quad Langfun & 71.5 & - & - \\
\quad Trase & 70.3 & - & - \\
\quad OWL & 69.1 & - & - \\
\hline \\
\multicolumn{1}{l}{\textit{Software Engineering Agents}} \\
\quad SWE-agent & - & 62.4 & - \\
\quad OpenHands & - & 65.8 & - \\
\hline \\
\multicolumn{1}{l}{\textit{Web Navigation Agents}} \\
\quad Browser use & - & - & 89.1 \\
\quad Operator & - & - & 87 \\
\quad Skyvern & - & - & 85.6 \\
\hline \\
\textbf{\aime{} (Ours)} & \textbf{77.6} & \textbf{66.4} & \textbf{92.3} \\
\end{tabular}
\end{center}
\end{table}

On \textbf{GAIA}, \aime{} achieves a new state-of-the-art success rate of 77.6\%, outperforming the strongest baselines like Langfun. We attribute this significant performance gain to the \textit{Dynamic Planner}, which allows the system to flexibly adapt its strategy when initial reasoning paths fail, a crucial capability for GAIA's complex, multi-step problems.

On \textbf{SWE-bench Verified}, \aime{} resolves 66.4\% of the issues, surpassing top specialized agents like OpenHands. While the performance of leading agents in this domain is highly competitive, we believe our advantage stems from the \textit{Actor Factory}. It can instantiate different types of agents on-the-fly (e.g., a "code-reader" to understand context, then a "debugger" to isolate the fault), leading to a more robust and effective problem-solving process.

On \textbf{WebVoyager}, \aime{} demonstrates superior robustness in live web environments, achieving an impressive 92.3\% success rate. This performance exceeds strong baselines like Browser use. Unlike agents with fixed plans that may falter with unexpected website changes, \aime{}'s tight feedback loop between the \textit{Dynamic Actors} and the \textit{Dynamic Planner} enables it to immediately re-plan and recover from errors, resulting in higher task completion.

\section{Related Work}
\label{sec:related}

The field of Multi-Agent Systems (MAS) has been transformed by the rise of Large Language Models (LLMs)~\citep{DBLP:conf/nips/BrownMRSKDNSSAA20, DBLP:journals/corr/abs-2303-08774}. Unlike traditional systems that relied on rigid, formal planning models~\citep{DBLP:books/daglib/0014222, DBLP:conf/icmas/RaoG95}, modern MAS leverage LLMs as cognitive engines, enabling unprecedented flexibility and coordination through natural language. This has spurred the development of new collaboration paradigms, which we review below.

\subsection{Role-Based Multi-Agent Collaboration}
A dominant paradigm in modern LLM-based MAS involves assigning specialized roles to agents to decompose complex tasks, often inspired by human organizational structures. Frameworks like \textbf{MetaGPT}~\citep{DBLP:conf/iclr/HongZCZCWZWYLZR24} and \textbf{ChatDev}~\citep{DBLP:conf/acl/QianLLCDL0CSCXL24} simulate a software company, where agents playing roles like "product manager" or "engineer" follow structured protocols to achieve their objectives. Similarly, \textbf{MAGIS}~\citep{DBLP:conf/nips/0003ZWZ0C24} and \textbf{MarsCode Agent}~\citep{DBLP:journals/corr/abs-2409-00899} design dedicated Standard Operating Procedures (SOPs) for software development. \textbf{CodeR}~\citep{DBLP:journals/corr/abs-2406-01304} extends this by predefining multiple SOPs and selecting one based on the task at hand. While these systems demonstrate the power of structured collaboration, their workflows and agent capabilities are largely static. This rigidity limits their ability to adapt to unforeseen circumstances or tasks that deviate from the predefined SOPs. 
Other frameworks like \textbf{AutoGen}~\citep{DBLP:journals/corr/abs-2308-08155} and \textbf{AgentVerse}~\citep{DBLP:journals/corr/abs-2308-10848} offer more flexible communication patterns, but the definition of agent roles and their capabilities often remains fixed.

\subsection{Automated Agent Architecture Design}
Recognizing the limitations of static designs, a recent line of research focuses on automatically searching for optimal agent architectures. These approaches, however, typically aim to find a superior \textit{static} design before execution begins.
\begin{itemize}
    \item \textit{Workflow Optimization:} Several works aim to automate the generation of the collaboration plan itself. For example, \textbf{AOP}~\citep{DBLP:conf/iclr/LiXLTDL25} investigates an agent-oriented planning method that leverages fast task decomposition and a reward model for efficient evaluation. Others, like \textbf{AFlow}~\citep{DBLP:conf/iclr/ZhangXYTCCZCHWZ25} and \textbf{Flow}~\citep{DBLP:conf/iclr/NiuSLS00L25}, automatically generate graph-based workflows, though often with the simplifying assumption of homogeneous agent capabilities. More advanced approaches such as \textbf{Agentic Supernet}~\citep{DBLP:journals/corr/abs-2502-04180} and \textbf{FlowReasoner}~\citep{DBLP:journals/corr/abs-2504-15257} even learn to generate these workflows from predefined agentic operators. A common limitation, however, is that these methods produce a static collaboration plan \textit{prior to execution}, making them vulnerable to real-time events that deviate from the initial strategy.

    \item \textit{Agent Role Optimization:} Complementary research focuses on optimizing individual agent design. \textbf{AgentSquare}~\citep{DBLP:conf/iclr/ShangLZMLXL25} searches for an optimal agent architecture by composing it from a set of given modules, while \textbf{ADAS}~\citep{DBLP:journals/corr/abs-2408-08435} automates the creation of a single agent via code generation. These methods enhance agent components but do not directly address the dynamics of multi-agent collaboration.

\end{itemize}

In contrast to these methods, \aime{} provides a framework for dynamic adaptation \textit{during} execution. It combines a \textit{Dynamic Planner}, which refines plans based on real-time outcomes, with an \textit{Actor Factory} that implements \textbf{Dynamic Actor Instantiation} to assemble specialized agents on-demand. This approach of combining reactive planning with on-the-fly specialization offers a practical and resilient solution for adaptability, avoiding the high computational overhead of offline architecture search.
\section{Conclusion}
\label{sec:conclusion}

This paper introduced \aime{}, a novel framework that enables dynamic, reactive collaboration through three key innovations: a \textit{Dynamic Planner} for adaptive strategy, an \textit{Actor Factory} for on-demand instantiation of specialized agents, and a centralized \textit{Progress Management Module} for coherent state awareness. It was designed specifically to address critical weaknesses in the conventional plan-and-execute framework, namely its rigid planning, static agent roles, and inefficient communication. Our experiments confirm that this approach is highly effective, with \aime{} significantly outperforming traditional models in adaptability, efficiency, and overall task success rate. 

Our future work focus on enhancing scalability for larger agent teams and empowering agents to autonomously acquire new capabilities, reducing their reliance on pre-curated tools. By shifting the paradigm from static execution to dynamic adaptation, \aime{} represents a significant step toward building more resilient and intelligent autonomous systems.

% \subsubsection*{Author Contributions}
% If you'd like to, you may include  a section for author contributions as is done
% in many journals. This is optional and at the discretion of the authors.

% \subsubsection*{Acknowledgments}
% Use unnumbered third level headings for the acknowledgments. All
% acknowledgments, including those to funding agencies, go at the end of the paper.

\bibliography{iclr2025_conference}
\bibliographystyle{iclr2025_conference}

% \appendix
% \input{7_appendix}

\end{document}